\documentclass[twocolumn]{article}

\usepackage{authblk}
\usepackage{amsmath}
\usepackage{amssymb}
\usepackage{mleftright}
\usepackage{graphicx}
\usepackage{svg}
\usepackage{pifont}
\newcommand{\cmark}{\ding{51}}
\newcommand{\xmark}{\ding{55}}
\usepackage{makecell}
\usepackage{colortbl}
\DeclareMathOperator*{\expect}{\mathbb{E}}
\usepackage{graphicx}
\usepackage{algorithm}
\usepackage[noend]{algpseudocode}

\usepackage{tikz}
\usetikzlibrary{positioning}
\usetikzlibrary{fit}
\usepackage[super]{natbib}
\usepackage{times}
\usepackage[page]{appendix}

\title{AI planning in the imagination: High-level planning on learned abstract search spaces}

\author[1]{Carlos Martin}
\author[1,2,3,4]{Tuomas Sandholm}

\affil[ ]{cgmartin@cs.cmu.edu, sandholm@cs.cmu.edu}
\affil[1]{Carnegie Mellon University}
\affil[2]{Strategy Robot, Inc.}
\affil[3]{Optimized Markets, Inc.}
\affil[4]{Strategic Machine, Inc.}

\date{}

\begin{document}

\maketitle
\begin{abstract}
Search and planning algorithms have been a cornerstone of artificial intelligence since the field's inception.
Giving reinforcement learning agents the ability to plan during execution time has resulted in significant performance improvements in various domains.
However, in real-world environments, the model with respect to which the agent plans has been constrained to be grounded in the real environment itself, as opposed to a more abstract model which allows for planning over compound actions and behaviors.
We propose a new method, called PiZero, that gives an agent the ability to plan in an abstract search space that the agent learns during training, which is completely decoupled from the real environment.
Unlike prior approaches, this enables the agent to perform high-level planning at arbitrary timescales and reason in terms of compound or temporally-extended actions, which can be useful in environments where large numbers of base-level micro-actions are needed to perform relevant macro-actions.
In addition, our method is more general than comparable prior methods because it seamlessly handles settings with continuous action spaces, combinatorial action spaces, and partial observability.
We evaluate our method on multiple domains, including the traveling salesman problem, Sokoban, 2048, the facility location problem, and Pacman.
Experimentally, it outperforms comparable prior methods without assuming access to an environment simulator at execution time.
\end{abstract}

\section{Introduction}

In many domains, planning at execution time significantly improves performance.
In games like chess and Go,
and single-agent domains like Sokoban, Pacman, and 2048,
all state-of-the-art approaches use planning.
No planning-free, reactive approach has matched their performance.

We present a new method that gives agents the ability to plan in a learned abstract search space that is completely decoupled from the real environment. Unlike prior approaches, this allows agents to perform \emph{high-level planning} at arbitrary timescales, and reason in terms of compound or temporally-extended actions, which can be useful in environments where many ``micro'' actions are needed to perform relevant ``macro'' actions.

Lookahead search and reasoning is a central paradigm in artificial intelligence.
Traditionally, the search space is first formulated and then solved \citep{newell1965search,hart1968formal,nilsson1971problem,hart1972correction,lanctot2017unified,Brown18:Depth}.
Here, the formulation of the search space happens during the solve.
This breaks the traditional paradigm because it does not require the search space to be formulated up front.

Our method augments a standard neural-network-based agent with a \emph{planning module} that it can query before selecting actions in the real environment. The planning module uses a learned \emph{dynamics model} and a learned \emph{prediction model}. The dynamics model takes a state and action as input and returns a reward and new state as output. The prediction model takes a state as input and returns an estimated abstract value and abstract action probabilities. The states, actions, and rewards of the planning module are purely abstract and do not need to have any direct relation or correspondence with those of the real environment. They need not have the same dimensionality or even type (discrete versus continuous).

In each step of an episode, the following occurs. First, the agent encodes its current memory and observation into an \emph{abstract state} for its planning module (where the encoding itself is learned). Second, the planning module performs Monte Carlo Tree Search (MCTS) in an \emph{abstract search space} created by a learned abstract dynamics model.
In our case, we assume the abstract action space is a discrete set of \(k\) actions, where \(k\) is a fixed hyperparameter. Third, the results of the MCTS subroutine are passed into another network which outputs an action for the real environment. Fourth, the observation and generated action are used to update the agent's memory.

In \S\ref{sec:related_work}, we describe prior algorithms in the literature and related research. 
In \S\ref{sec:background}, we formulate the problem we are tackling in detail.
In \S\ref{sec:method}, we motivate and describe our method.
In \S\ref{sec:experiments}, we present our experiments.
In \S\ref{sec:conclusion}, we present conclusions and future research.

\section{Background}
\label{sec:background}

We use the following notation throughout the paper: \(\triangle \mathcal{X}\) is the set of probability distributions on \(\mathcal{X}\), \([n] = \{0, \ldots, n - 1\}\), \(a \mathop{\|} b\) is the concatenation of vectors \(a\) and \(b\), \(\dim \mathcal{X}\) is the dimensionality of a space \(\mathcal{X}\), and \(|\mathcal{X}|\) is the cardinality of a set \(\mathcal{X}\).

A \emph{finite-horizon partially-observable Markov decision process} (FH-POMDP), henceforth called an \emph{environment}, is a tuple \((\mathcal{S}, \mathcal{O}, \mathcal{A}, \rho, \delta, H)\) where 
\(\mathcal{S}\) is a state space,
\(\mathcal{O}\) is an observation space,
\(\mathcal{A}\) is an action space,
\(\rho : \triangle \mathcal{S}\) is an initial state distribution, 
\(o : \mathcal{S} \to \mathcal{O}\) is an observation function,
\(\delta : \mathcal{S} \times \mathcal{A} \to \triangle (\mathcal{S} \times \mathbb{R})\) is a transition function,
and \(H \in \mathbb{N}\) is a horizon.

An \emph{agent} is a tuple \((\mathcal{M}, m_0, \pi)\) where 
\(\mathcal{M}\) is a memory space,
\(m_0 \in \mathcal{M}\) is an initial memory,
and \(\pi : \mathcal{O} \times \mathcal{M} \to \triangle (\mathcal{A} \times \mathcal{M})\) is a transition function. 
The agent's memory can implicitly encode the history of past observations and actions.

An episode or \emph{trajectory} \(\tau = (m, s, o, a, r)\) is a tuple of sequences of memories, states, actions, and rewards.
A trajectory is generated as follows.
First, an initial state \(s_0 \sim \rho\) is sampled.
Then, the agent and environment are updated in tandem for all \(t \in [H]\):
The agent is updated with environment's observation, \(a_t, m_{t+1} \sim \pi(o(s_t), m_t)\),  and the environment is updated with agent's action, \(s_{t+1}, r_t \sim \delta(s_t, a_t)\). The \emph{score} or \emph{return} of a trajectory is the sum of its rewards, \(R = \sum_{t \in [H]} r_t\). Given an environment, our goal is to obtain an agent that maximizes the expected return \(\expect_\tau R\).

\section{Proposed method}
\label{sec:method}

In this section, we describe our proposed method.

\subsection{Agent architecture}

The core architecture of the agent is illustrated below.

\begin{center}
\begin{tikzpicture}
\node (cstate) {state};
\node [right=60pt of cstate] (astate) {state};
\node [below=20pt of astate] (aaction) {action};
\node [below=20pt of cstate] (caction) {action};
\draw [->] (cstate) -- node [above] {encoder} (astate);
\draw [->] (astate) -- node [right] (planner) {planner} (aaction);
\draw [->] (aaction) -- node [below] {decoder} (caction);
\draw [->,dotted] (cstate) -- node [left] (policy) {policy} (caction);
\node [fit=(cstate)(caction)(policy),draw,dotted,rounded corners,inner sep=6pt] (c) {};
\node [fit=(astate)(aaction)(planner),draw,dotted,rounded corners,inner sep=6pt] (a) {};
\node [below=0pt of c] {concrete};
\node [below=0pt of a] {abstract};
\end{tikzpicture}
\end{center}

The agent's current memory and observation constitute the agent's \emph{concrete state}.
The concrete state is passed to an \emph{encoder}, which transforms it into an \emph{abstract state}.
This is a state in an abstract space whose dynamics is learned by the agent.
The abstract state is passed to a \emph{planner}, which performs planning in the abstract space and returns an \emph{abstract action}.
The abstract action is passed to a \emph{decoder}, which transforms it into a \emph{concrete action} for the real environment.\footnote{If stochastic actions are desired, the abstract action can be concatenated with \emph{latent noise}, e.g. from a standard normal distribution, before being passed to the decoder. There are situations where randomizing over actions is beneficial, \emph{e.g.}, when the agent has imperfect recall or the environment is adversarial (\emph{e.g.}, due to the presence of other agents). The approach of feeding noise as input to induce a stochastic output is called an implicit density model or deep generative model in the literature~\citep{Ruthotto_2021}, and has the advantage of creating flexible classes of distributions. For example, sampling from a categorical distribution is equivalent to perturbing its logits with Gumbel noise and applying argmax.}
The agent's memory is then updated with the resulting action and input observation.

For the agent's memory, we use a recurrent unit.
Many recurrent units have been proposed in the literature~\citep{Salehinejad_2017,Yu_2019}.
Among the most popular are Long Short-Term Memory (LSTM)~\citep{Hochreiter_1997} and Gated Recurrent Unit (GRU)~\citep{Cho_2014a,Cho_2014b}.
We use GRU since it is simpler, faster, and has comparable performance to LSTM~\citep{Jozefowicz_2015,Khandelwal_2016,Yang_2020}.
For the encoder and decoder, we use standard feedforward networks.

\subsection{Planner architecture}

Our planner performs MCTS in an abstract space learned by the agent.
More precisely, let \(\mathcal{S}\) be a state space and \(\mathcal{A}\) be an action space (we use \(\mathcal{A} = [n]\) for some \(n \in \mathbb{N}\)).
The dynamics function \(\delta : \mathcal{S} \times \mathcal{A} \to \mathbb{R} \times \mathcal{S}\) maps a state and action to a reward and new state.
The prediction function \(\phi : \mathcal{S} \to \mathbb{R} \times \mathbb{R}^\mathcal{A}\) maps a state to a predicted value and action logits.
The parameters of both of these functions are learned.
The planner runs an MCTS algorithm (described in the next section) on the resulting space, starting at the abstract state fed into the planner, and outputs the abstract action returned by the MCTS algorithm.

For \(\delta\), we use a GRU and apply a one-hot encoding to its input action.
For \(\phi\), we use a standard feedforward network.
For the MCTS algorithm, we use the open-source JAX~\citep{jax} library called mctx~\citep{deepmind2020jax}.

This architecture is inspired by MuZero~\citep{muzero}, but there are some important differences that we describe in \S\ref{sec:related_work}.

\subsection{Planning algorithm}

PiZero uses variant of MCTS introduced by \citet{alphazero}.
This algorithm iteratively constructs a search tree starting from some given state \(s_0\).
Each node in this search tree contains a state \(s\) and, for each action \(a\) from \(s\), the visit count \(N(s, a)\), value estimate \(Q(s, a)\), prior probability \(P(s, a)\), reward \(R(s, a)\), and successor state \(S(s, a)\).
Each iteration of the algorithm consists of 3 phases: \emph{selection}, \emph{expansion}, and \emph{backpropagation}.

\paragraph{Selection}
The tree is traversed starting from the root node until a leaf edge is reached.
At each internal node s the algorithm selects the action \(a\) which maximizes the upper confidence bound proposed in \citet{alphago} and shown in Equation \ref{eq:ucb}.
\begin{align} \label{eq:ucb}
    Q(s, a) + P(s, a) \frac{\sqrt{1 + \sum_b N(s, b)}}{1 + N(s, a)} \times \\
    \mleft( c_1 + \log \frac{\sum_b N(s, b) + c_2 + 1}{c_2} \mright)
\end{align}
Here, \(c_1, c_2\) are constants that control the relative importance of the value estimates and prior probabilities. We use the same values as s\citet{stochastic_muzero}.
After \(a\) is selected, the reward \(r = R(s, a)\) and successor state \(s' = S(s, a)\) are queried from the node's lookup table, and the successor state's node becomes the new node.

\paragraph{Expansion}
When a leaf edge is reached, the reward and successor state \((r, s') = g_\theta(s, a)\) are computed from the dynamics function and stored in the node's lookup table.
The policy and value are computed by the prediction function \((p, v) = f_\theta(s)\).
A new node is created corresponding to \(s'\) and added to the search tree. The lookup table is initialized, for each action \(a\), with \(N(s, a) = 0, Q(s, a) = 0, P(s, a) = p\).

\paragraph{Backpropagation}
The value estimate of the newly added edge is backpropagated up the tree along the trajectory from the root to the leaf using an \(n\)-step return estimate.
Specifically, from \(t = T\) to \(0\), where \(T\) is the length of the trajectory, we compute a \(T-t\)-step estimate of the cumulative discounted reward that bootstraps from the value function \(v_l\):
\(
    G_t = \sum_{t=0}^{l-1-t} \gamma^t r_{t+1+t} + \gamma^{T-t} v_l
\). For each such \(t\), we also update the statistics for the edge \((s_t, a_t)\) as follows:
\(
    Q(s_t, a_t) := \frac{N(s_t, a_t) Q(s_t, a_t) + G_t}{N(s_t, a_t) + 1}
\),
\(
    N(s_t, a_t) := N(s_t, a_t) + 1
\). This algorithm can be extended to include \emph{chance nodes} in the search tree, as described in \citet{stochastic_muzero}.
The action probabilities at a chance node are output by a learned function.

\subsection{Training procedure}

Once we have fixed an architecture for the agent, we seek to find parameters that maximize its expected return in the environment.
One way to do this is by using stochastic gradient ascent. Stochastic gradient ascent requires an estimator of the gradient of the expected return with respect to the parameters. This is called the policy gradient. There is a vast literature on policy gradient methods, including REINFORCE \citep{reinforce} and actor-critic methods \citep{actor_critic_1,actor_critic_2}.

However, there is a problem: Most of these methods assume that the policy is \emph{differentiable}---that is, that its output (an action distribution) is differentiable with respect to the parameters of the policy. However, our planning policy uses MCTS as a subroutine, and standard MCTS is not differentiable. Because our policy contains a non-differentiable submodule, we need to find an alternative way to optimize the policy's parameters.

Fortunately, we can turn to black-box (\emph{i.e.}, zeroth-order) optimization. Black-box optimization uses only function evaluations to optimize a black-box function with respect to a set of inputs.
In particular, it does not require gradients.
In our case, the black-box function maps our policy's parameters to a sampled episode score.

There is a class of black-box optimization algorithms called evolution strategies (ES)~\citep{Rechenberg_1973, Schwefel_1977, Rechenberg_1978} that maintain and evolve a population of parameter vectors. Natural evolution strategies (NES)~\citep{Wierstra_2008, Wierstra_2014, Sun_2009} represent the population as a distribution over parameters and maximize its average objective value using the score function estimator.

For many parameter distributions, such as Gaussian smoothing, this is equivalent to evaluating the function at randomly-sampled points and estimating the gradient as a sum of estimates of directional derivatives along random directions \citep{Duchi_2015, Nesterov_2017, Shamir_2017, Berahas_2022}.

ES has been shown to be a scalable alternative to standard reinforcement learning \citep{openai_es}.
\citet{lenc2019non} show that ES is also a viable method for learning non-differentiable parameters of large supervised models.
We use OpenAI-ES~\citep{openai_es}, an NES algorithm that is based on the identity
\(
    \nabla_\mathbf{x} \expect_{\mathbf{z} \sim \mathcal{N}} f(\mathbf{x} + \sigma \mathbf{z}) = \tfrac{1}{\sigma} \expect_{\mathbf{z} \sim \mathcal{N}} f(\mathbf{x} + \sigma \mathbf{z}) \mathbf{z}
\),
where \(\mathcal{N}\) is the standard multivariate normal distribution with the same dimension as \(\mathbf{x}\).
The algorithm computes works as follows.
Let \(\mathcal{I}\) be a set of indices.
For each \(i \in \mathcal{I}\) in parallel, sample \(\mathbf{z}_i \sim \mathcal{N}\) and compute \(\delta_i = f(\mathbf{x} + \sigma \mathbf{z}_i)\).
Finally, compute the pseudogradient \(\mathbf{g} = \frac{1}{\sigma |\mathcal{I}|} \sum_{i \in \mathcal{I}} \delta_i \mathbf{z}_i\).
To reduce variance, like \citet{openai_es}, we use antithetic sampling~\citep{Geweke_1988}, also called mirrored sampling~\citep{brockhoff2010mirrored}, which uses pairs of perturbations \(\pm \sigma \mathbf{z}_i\).
The resulting gradient estimate is fed into a standard optimizer. For the latter, we use Adaptive Momentum (Adam)~\citep{adam}.

OpenAI-ES is massively parallelizable. Each \(\delta_i\) can be evaluated on a separate worker. Furthermore, the entire optimization procedure can be performed with minimal communication bandwidth between workers. All workers are initialized with the same random seed. Worker \(i\) evaluates \(\delta_i\), sends it to the remaining workers, and receives the other workers' values (this is called an allgather operation in distributed computing). Thus the workers compute the same \(\mathbf{g}\) and stay synchronized.
This process is described in Algorithm 2 of \citet{openai_es}.
The full training process is summarized in Algorithm 1.

\algnewcommand\algorithmicinput{\textbf{Input:}}
\algnewcommand\Input{\item[\algorithmicinput]}
\begin{algorithm}[t]
\caption{
Distributed training.
\\
The following algorithm runs on each worker.
All workers are initialized with the same random seed.
\\
The function \(f\) samples an episode with the input parameters for the agent and outputs its score.
}
\label{alg:algorithm}
\begin{algorithmic}
\State \(\mathbf{x} \gets \text{agent.initialize\_params}()\)
\State \(S \gets \text{optimizer.initialize\_state}(\mathbf{x})\)
\Loop
\State \(\mathbf{x} \gets \text{optimizer.get\_params}(S)\)
\State \(\mathcal{I} \gets\) set of available workers
\For{\(i \in \mathcal{I}\)}
\State \(\mathbf{z}_i \sim \mathcal{N}\)
\EndFor
\State \(j \gets \) own worker rank
\State \(\delta_{j} \gets f(\mathbf{x} + \sigma \mathbf{z}_{j}) - f(\mathbf{x} - \sigma \mathbf{z}_{j})\)
\State send \(\delta_{j}\) to other workers
\State receive \(\{\delta_i\}_{i \in \mathcal{I} - \{j\}}\) from other workers
\State \(\mathbf{g} \gets \frac{1}{2 \sigma |\mathcal{I}|} \sum_{i \in \mathcal{I}} \delta_i \mathbf{z}_i\)
\State \(S \gets \text{optimizer.update\_state}(S, \mathbf{g})\)
\EndLoop
\end{algorithmic}
\end{algorithm}

\section{Related work}
\label{sec:related_work}

Monte-Carlo evaluation estimates the value of a position by averaging the return of several random rollouts.
It can serve as an evaluation function for the leaves of a search tree.
Monte-Carlo Tree Search (MCTS) \citep{mcts_original} combines Monte-Carlo evaluation with tree search.
Instead of backing-up the min-max value close to the root, and the average value at some depth, a more general backup operator is defined that progressively changes from averaging to min-max as the number of simulations grows.
This provides a fine-grained control of the tree growth and allows efficient selectivity.
\citet{mcts_review} present a survey of recent modifications and applications of MCTS.

AlphaGo \citep{alphago} used a variant of MCTS to tackle the game of Go.
This variant uses a neural network to evaluate board positions \emph{and} select moves.
These networks are trained using a combination of supervised learning from human expert games and reinforcement learning from self-play.
It was the first computer program to defeat a human professional player.

AlphaGo Zero \citep{alphago_zero} used reinforcement learning alone, \emph{without} any human data, guidance or domain knowledge beyond game rules.
AlphaZero \citep{alphazero} generalized AlphaGo Zero into a single algorithm that achieved superhuman performance in many challenging games, including chess and shogi.

MuZero~\citep{muzero} combined AlphaZero's tree-based search with a \emph{learned dynamics model}. The latter allows it to plan in environments where the agent does \emph{not} have access to a simulator of the environment at execution time.
In the authors' words, ``All parameters of the model are trained jointly to accurately match the policy, value, and reward, for every hypothetical step \(k\), to corresponding target values observed after \(k\) actual time-steps have elapsed.''
In other words, unlike our method, its dynamics and prediction models are coupled to the real environment.
Furthermore, since the action probabilities returned by the MCTS process are treated as fixed targets to train the prediction model, MuZero does not need to differentiate through the nondifferentiable MCTS process.

\citet{hubert2021learning} proposed Sampled MuZero, an extension of the MuZero algorithm that is able to learn in domains with arbitrarily complex action spaces (including ones that are continuous and high-dimensional) by planning over sampled actions.

Stochastic MuZero~\citep{stochastic_muzero} extended MuZero to environments that are inherently stochastic, partially observed, or so large and complex that they appear stochastic to a finite agent.
It learns a stochastic model incorporating afterstates, and uses this model to perform a stochastic tree search.
It matches the performance of MuZero in Go while matching or exceeding the state of the art in a set of canonical single and multiagent environments, including 2048 and backgammon.

Our method has some important differences to prior methods.
While some prior methods use a learned dynamics model, like ours, their training objective is different. In particular, they train a model to predict accurate rewards and values \emph{for the real environment}. In contrast, our method \emph{completely decouples} the dynamics and evaluation models from the real environment. This gives agents the ability to perform arbitrarily \emph{high-level planning} in a learned \emph{abstract search space} whose rewards and values are not tied to those of the real environment in any way.

Since neither states, actions, rewards, nor values in the abstract space are tied to the timescale of the environment, the agent has the ability to reason in terms of compound, temporally-extended actions or behaviors.
This is useful in environments, such as real-time strategy (RTS) games, where it takes a very large number of ``micro-actions'' to perform ``macro-actions'' that are relevant to planning, such as reaching a target across the map and destroying it. Because they are so fine-grained, performing MCTS on the environment's real ``micro-actions'' would not yield useful information for any reasonable simulation budget.

Another difference to prior methods is that ours can straightforwardly handle continuous-action environments \emph{and} does not require computing probabilities of actions. Computing action probabilities for continuous action distributions requires constraining to a model class such as a parametric distribution or a normalizing flow~\citep{Kobyzev_2021,Papamakarios_2021}.

In terms of the optimization objective, a crucial difference between PiZero and prior methods like AlphaZero and MuZero is that, while the latter try to minimize a \emph{planning loss} for the value and policy networks, ours simply tries to \emph{directly maximize the episode score} in any way it can, which more directly corresponds to the overarching goal of reinforcement learning and brings advantages.

For example, our method can straightforwardly handle partially-observable environments where the agent must also \emph{learn what to remember} from observations, as opposed to having access to some \emph{fixed} representation of the observation history that is not learned. Since PiZero learns to maximize the episode score in an end-to-end fashion, it can learn what to remember without requiring any separate loss or objective function.

Table~\ref{tbl:comparison} summarizes some of the differences between our method and prior ones.
Additional related work can be found in the appendix.

\newcommand{\T}{\cellcolor{green!25}\cmark}
\newcommand{\F}{\cellcolor{red!25}\xmark}

\begin{table*}[t]
\centering
\begin{tabular}{|l|lllll|}
    \hline
    & AlphaZero & MuZero & Sampled MuZero & Stochastic MuZero & PiZero \\
    \hline
    No simulator & \F & \T & \T & \T & \T \\
    Stochastic model & \F & \F & \F & \T & \T \\
    Continuous actions & \F & \F & \T & \F & \T \\
    No observation storage & \F & \F & \F & \F & \T \\
    \makecell[l]{No action probabilities} & \F & \F & \F & \F & \T \\
    \makecell[l]{Low bandwidth} & \F & \F & \F & \F & \T \\
    \makecell[l]{No episode storage} & \F & \F & \F & \F & \T \\
    \makecell[l]{Backprop-free} & \F & \F & \F & \F & \T \\
    \hline
\end{tabular}
\caption{Comparison to prior methods.}
\label{tbl:comparison}
\end{table*}

\section{Experiments}
\label{sec:experiments}

We evaluate our method on multiple domains.
We note that we are not trying to develop the best special-purpose solver for any one of these benchmarks. Rather, our goal is to create a \emph{general} agent that can tackle a wide range of environments.
Unless otherwise noted, our experimental hyperparameters are those listed in Table~\ref{tab:hyperparameters}.
In our plots, solid lines indicate the mean across trials. Bands indicate a confidence interval for this mean with a confidence level of 0.95. The latter is computed using bootstrapping~\citep{Efron_1979}, specifically the bias-corrected and accelerated (BCa) method~\citep{Efron_1987}.

\subsection{Traveling salesman problem}

The traveling salesman problem (TSP) is a classic problem in combinatorial optimization. Given a set of cities and their pairwise distances, the goal is to find a shortest route that visits each city once and returns to the starting city. This problem has important applications in operations research, including logistics, computer wiring, vehicle routing, and various other planning problems~\citep{matai2010traveling}. TSP is known to be NP-hard~\citep{karp2010reducibility}.
Various approximation algorithms and heuristics~\citep{nilsson2003heuristics} have been developed for it.

Our environment is as follows. We seek to learn to solve TSP in general, not just one particular instance of it. Thus, on every episode, a new problem instance is generated by sampling a matrix \(M \sim \operatorname{Uniform}([0, 1]^{n \times 2})\), representing a sequence of \(n\) cities (in our experiments, we use 10 cities). At timestep \(t \in [n]\), the agent chooses a city \(a \in [n]\) to swap with the city in the \(t\)th position (possibly the same city, which leaves the sequence unchanged). At the end of the episode, the length of the tour through this sequence of cities (including the segment from the final city to the initial one) is computed, and treated as the \emph{negative} score. Thus the agent is incentivized to find the shortest tour through all the cities.
The agent observes the flattened matrix \(M\) together with the current timestep.

An example state is shown in Figure~\ref{fig:states}. Dots are cities, the filled dot is the initial city, and lines constitute the path constructed so far.
Results are shown in Figure~\ref{fig:results}, with equal run-time for each method. PiZero outperforms AlphaZero, despite lacking access to an environment simulator.

\subsection{Sokoban}

Sokoban is a puzzle in which an agent pushes boxes around a warehouse to get them to storage locations. It is played on a grid of tiles. Each tile may be a floor or a wall, and may contain a box or the agent. Some floor tiles are marked as storage locations. The agent can move horizontally or vertically onto empty tiles. The agent can also move a box by walking up to it and push it to the tile beyond, if the latter is empty.
Boxes cannot be pulled, and they cannot be pushed to squares with walls or other boxes. The number of boxes equals the number of storage locations. The puzzle is solved when all boxes are placed at storage locations.
In this puzzle, planning ahead is crucial since an agent can easily get stuck if it makes the wrong move.

Sokoban has been studied in the field of computational complexity and shown to be PSPACE-complete~\citep{sokoban_pspace_complete}. It has received significant interest in artificial intelligence research because of its relevance to automated planning (\emph{e.g.}, for autonomous robots), and is used as a benchmark. Sokoban's large branching factor and search tree depth contribute to its difficulty.
Skilled human players rely mostly on heuristics and can quickly discard several futile or redundant lines of play by recognizing patterns and subgoals, narrowing down the search significantly.
Various automatic solvers have been developed in the literature~\citep{junghanns1997sokoban,junghanns2001sokoban,froleyks2016using,shoham2020fess}, many of which rely on heuristics, but more complex Sokoban levels remain a challenge.

Our environment is as follows. We use the unfiltered Boxoban training set~\citep{boxobanlevels}, which contains levels of size \(10 \times 10\). At the beginning of each episode, a level is sampled from this dataset. The agent has four actions available to it, for motion in each of the four cardinal directions. The level ends after a given number of timesteps \(H \in \mathbb{N}\) (we use \(H = 50\)). The return at the end of an episode is the number of goals that are covered with boxes. Thus the agent is incentivized to cover all of the goals.

An example state is shown in Figure~\ref{fig:states}.
This image was rendered by JSoko \citep{jsoko}, an open-source Sokoban implementation.
The yellow vehicle is the agent, who must push the brown boxes into the goal squares marked with Xs.
(Boxes tagged ``OK'' are on top of goal squares.)
Results are shown in Figure~\ref{fig:results}, with equal run-time for each method.
PiZero outperforms AlphaZero, despite lacking access to an environment simulator.

\subsection{Collection problem}

In this environment, an agent must navigate an \(8 \times 8\) 2D gridworld to collect as many coins as possible within a time limit. At the beginning of each episode, coins are placed uniformly at random, as is the agent. On each timestep, the agent can move up, down, left, or right. The agent collects a coin when it moves into the same tile as the latter. After 20 timesteps, the score is minus the number of remaining coins. Thus the agent is incentivized to collect as many coins as possible within the time limit. A variant of this environment was used by \citet[\S 4.2]{vpn} as a benchmark.

This problem resembles a traveling salesman-like problem in which several ``micro'' actions are required to perform the ``macro'' actions of moving from one city to another. (Also, the agent can visit cities multiple times and does not need to return to its starting city.) This models situations where several fine-grained actions are required to perform relevant tasks, such as moving a unit in a real-time strategy game a large distance across the map. In order to plan effectively, an agent should be able reason in terms of such compound ``macro'' actions in some imagined abstract space.

An example state is shown in Figure~\ref{fig:states}. Brick tiles are walls, yellow circles are coins, and the human figure is the agent.
Results are shown in Figure~\ref{fig:results}, with equal run-time for each method. PiZero outperforms even AlphaZero, despite lacking access to an environment simulator.

\subsection{2048 game}

We use the 2048 implementation of Jumanji~\citep{bonnet2023jumanji}, a library of reinforcement learning environments written in JAX~\citep{jax}. Quoting from their documentation: ``2048 is a popular single-player puzzle game that is played on a 4x4 grid. The game board consists of cells, each containing a power of 2, and the objective is to reach a score of at least 2048 by merging cells together. The player can shift the entire grid in one of the four directions (up, down, right, left) to combine cells of the same value. When two adjacent cells have the same value, they merge into a single cell with a value equal to the sum of the two cells. The game ends when the player is no longer able to make any further moves. The ultimate goal is to achieve the highest-valued tile possible, with the hope of surpassing 2048. With each move, the player must carefully plan and strategize to reach the highest score possible.''
An example state is shown in Figure~\ref{fig:states}.
Results are shown in Figure~\ref{fig:results}.

\subsection{Facility location problem}

The facility location problem (FLP) is as follows. Given a set of clients, output a set of \(m\) facilities that minimizes the maximum (or mean) distance between a client and its closest facility.
This is a well-studied problem in location analysis, which is a branch of operations research and computational geometry that studies the optimal placement of facilities to minimize transportation costs.
It it also used for cluster analysis.
Computing an exact solution to this problem is NP-hard \citep{fowler1981optimal}.

Formally, let \(n \in \mathbb{N}\) be the number of clients, \(m \in \mathbb{N}\) be the number of facilities, and \(d : \mathbb{R}^2 \times \mathbb{R}^2 \to \mathbb{R}\) be the Euclidean distance. Then, given a set of clients \(C \in \mathbb{R}^{n \times 2}\), the goal is to minimize \(\max_{i \in n} \min_{j \in m} d(C_i, F_j)\) with respect to \(F \in \mathbb{R}^{m \times 2}\).
(In the mean-variant of this problem, the maximum is replaced with a mean.)
At the beginning of each episode, we randomly sample \(n\) points from the unit square \([0, 1]^2\), which are the clients.
On each step, the agent chooses a point on the unit square \([0, 1]^2\), which is a new facility to be placed.
The final score is the max-min metric described above.
Unlike previous environments, this environment has a \emph{continuous} action space.

An example state is shown in Figure \ref{fig:states}. Red dots are clients, blue dots are facilities placed so far, and black lines connect clients to their closest facility.
Results are shown in Figure~\ref{fig:results}.

\subsection{Pacman}

\begin{figure}
    \centering
    \frame{\includegraphics[width=.5\linewidth]{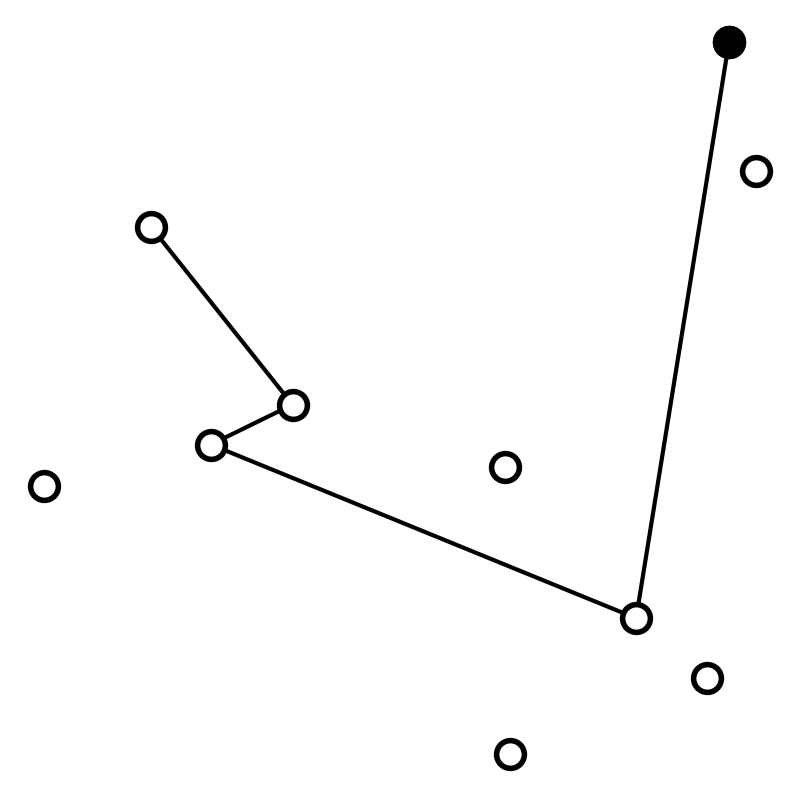}}%
    \frame{\includegraphics[width=.5\linewidth]{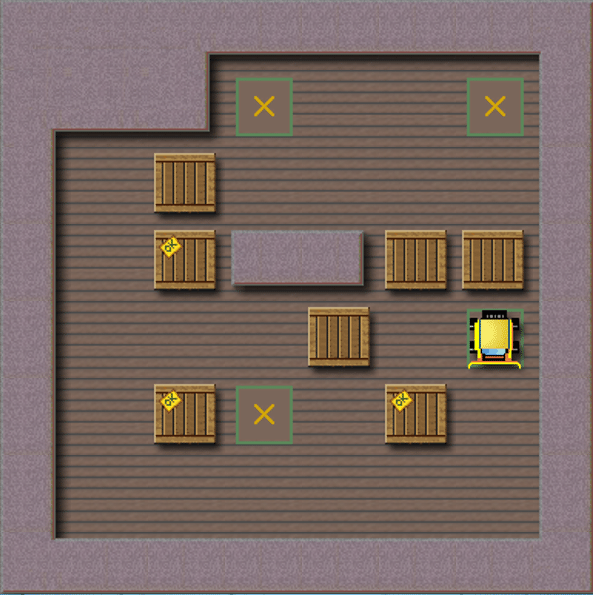}}
    \frame{\includegraphics[width=.5\linewidth]{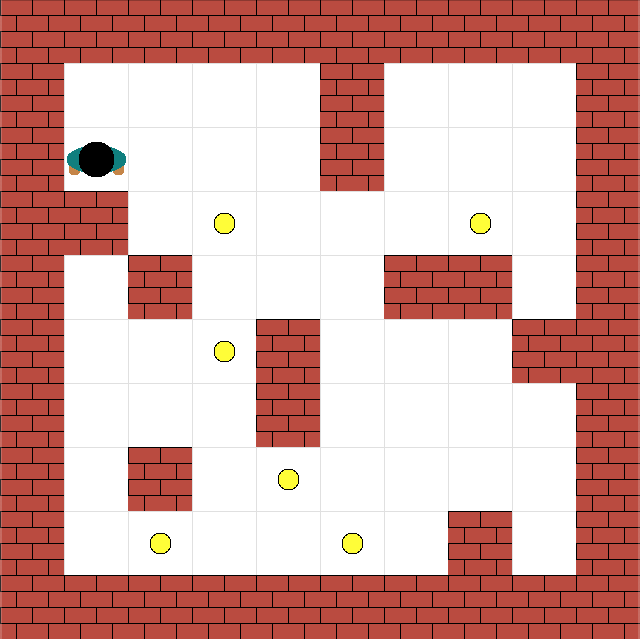}}%
    \frame{\includegraphics[width=.5\linewidth]{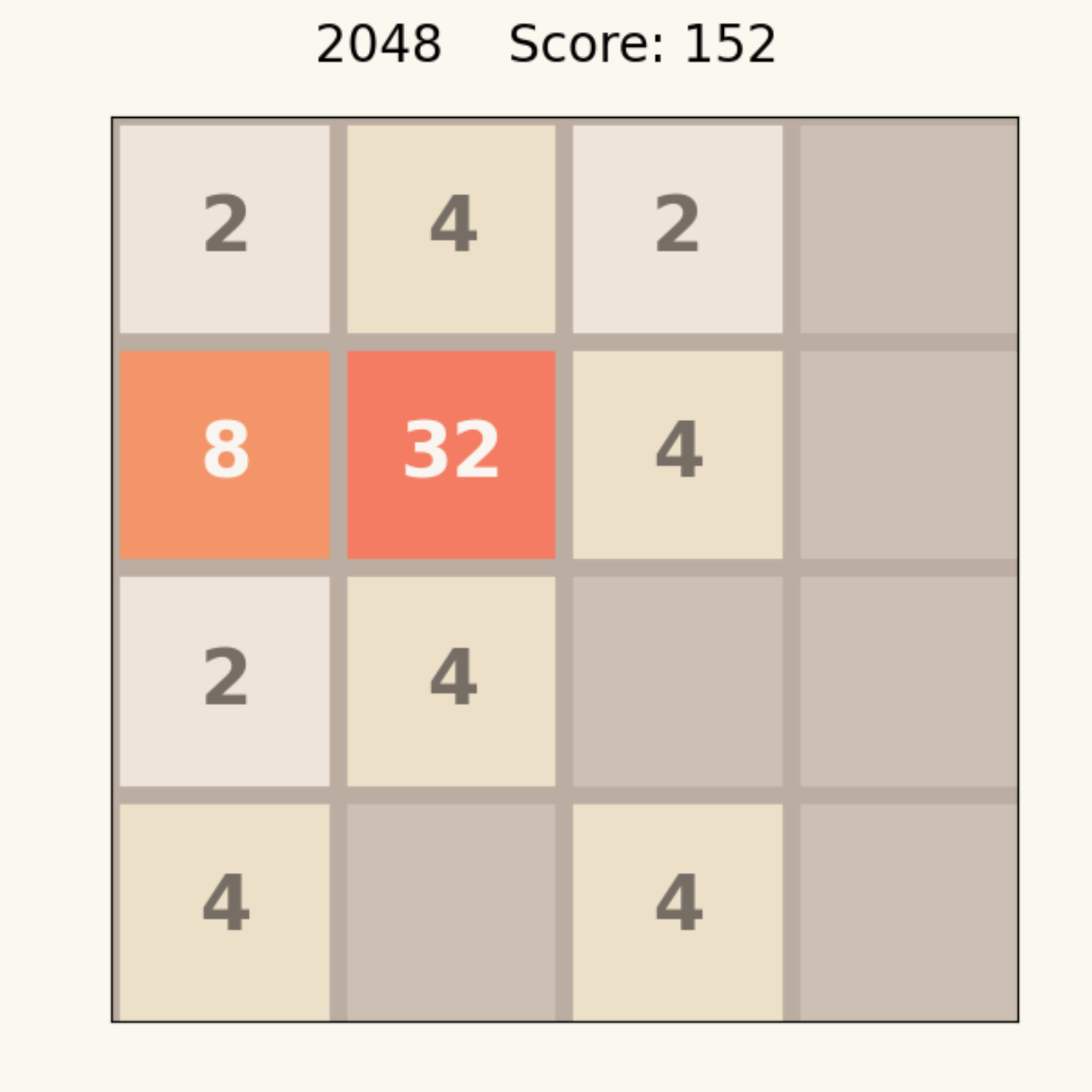}}
    \frame{\includegraphics[width=.5\linewidth]{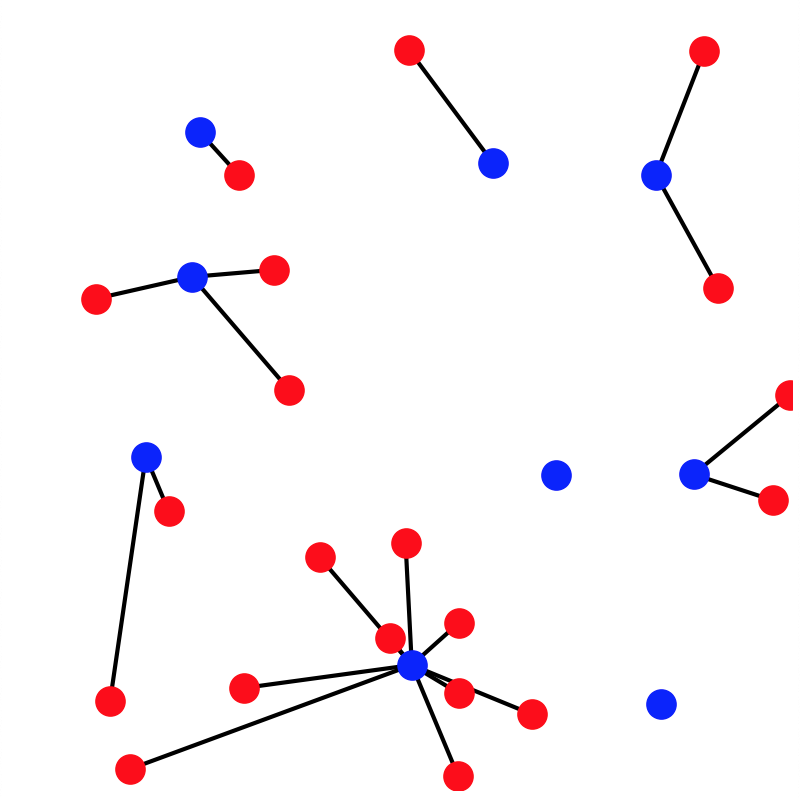}}%
    \frame{\includegraphics[width=.5\linewidth]{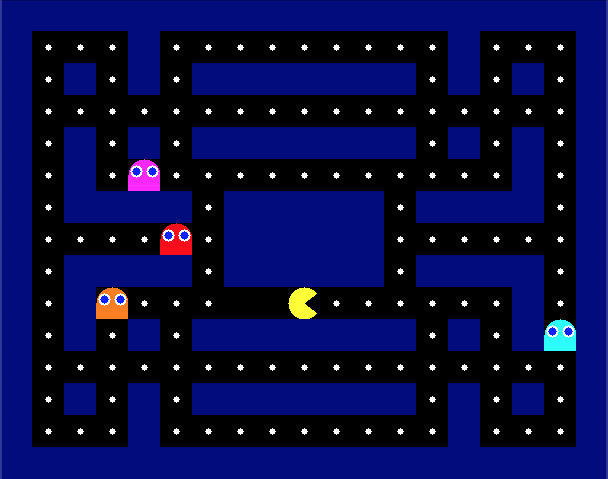}}
    \caption{States for TSP, Sokoban, Collect, 2048, FLP, and Pacman.}
    \label{fig:states}
\end{figure}

This is a simplified version of the classic video game called Pacman.
An example state is shown in Figure \ref{fig:states}.
The yellow circle is the player, which moves around a maze trying to eat as many dots as possible, while avoiding four colored ghosts that also move around the maze.

We also experiment on a \emph{partially-observable} version of Pacman where, at each step, the player can only see a fragment of the maze centered at their current position.
In this situation, it is important for the agent to be able to \emph{remember} what it has seen before (\emph{e.g.}, where they last saw each ghost).
To give the agent memory, we use a recurrent cell with a memory vector that is updated with the observation and action of the player at each step of the episode.
Since our method is able to train the \emph{entire} agent architecture in an end-to-end fashion, the agent can \emph{learn what to remember} from previous timesteps.

\begin{figure*}
    \centering
    \includegraphics[width=.333\linewidth]{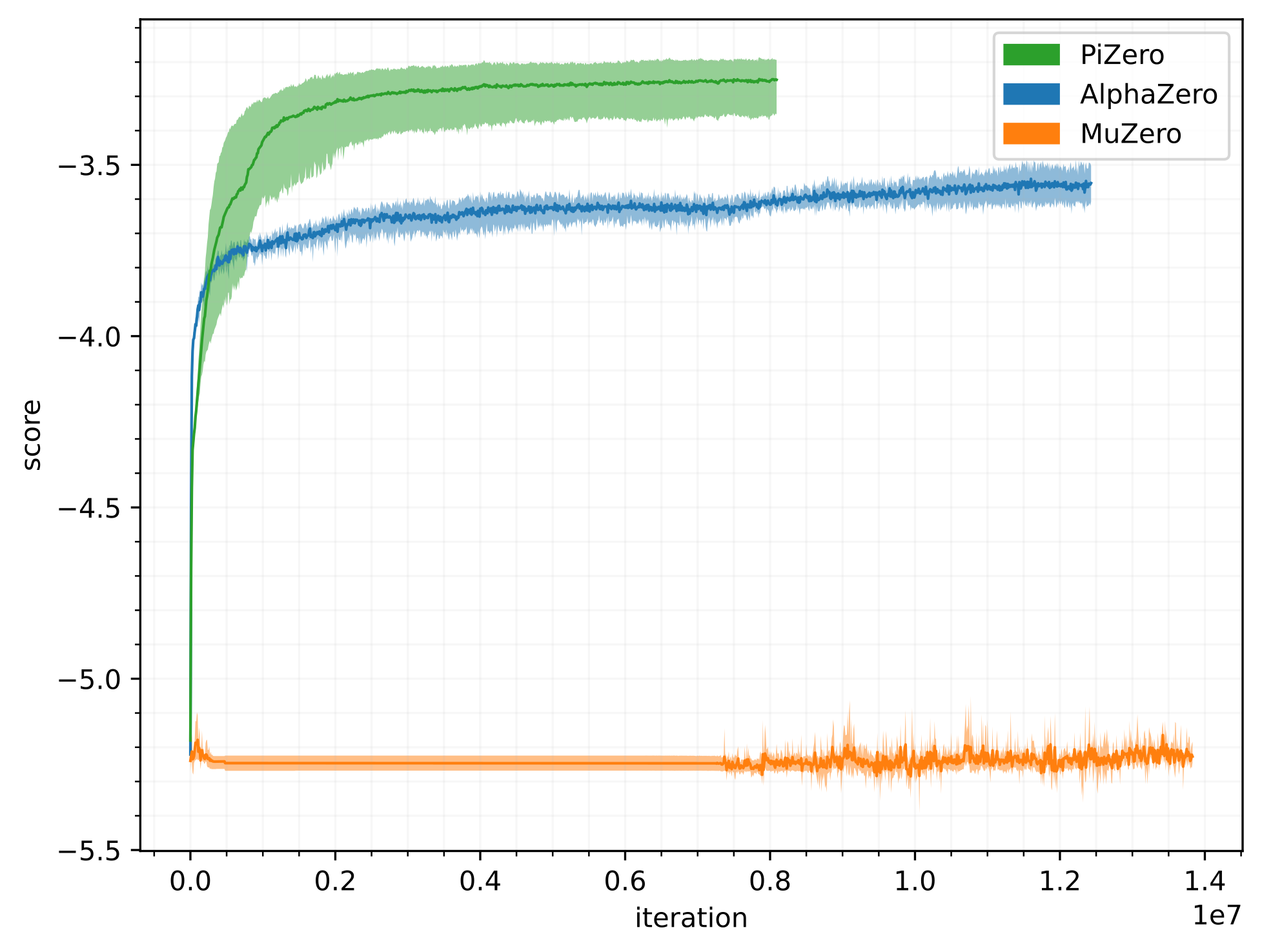}%
    \includegraphics[width=.333\linewidth]{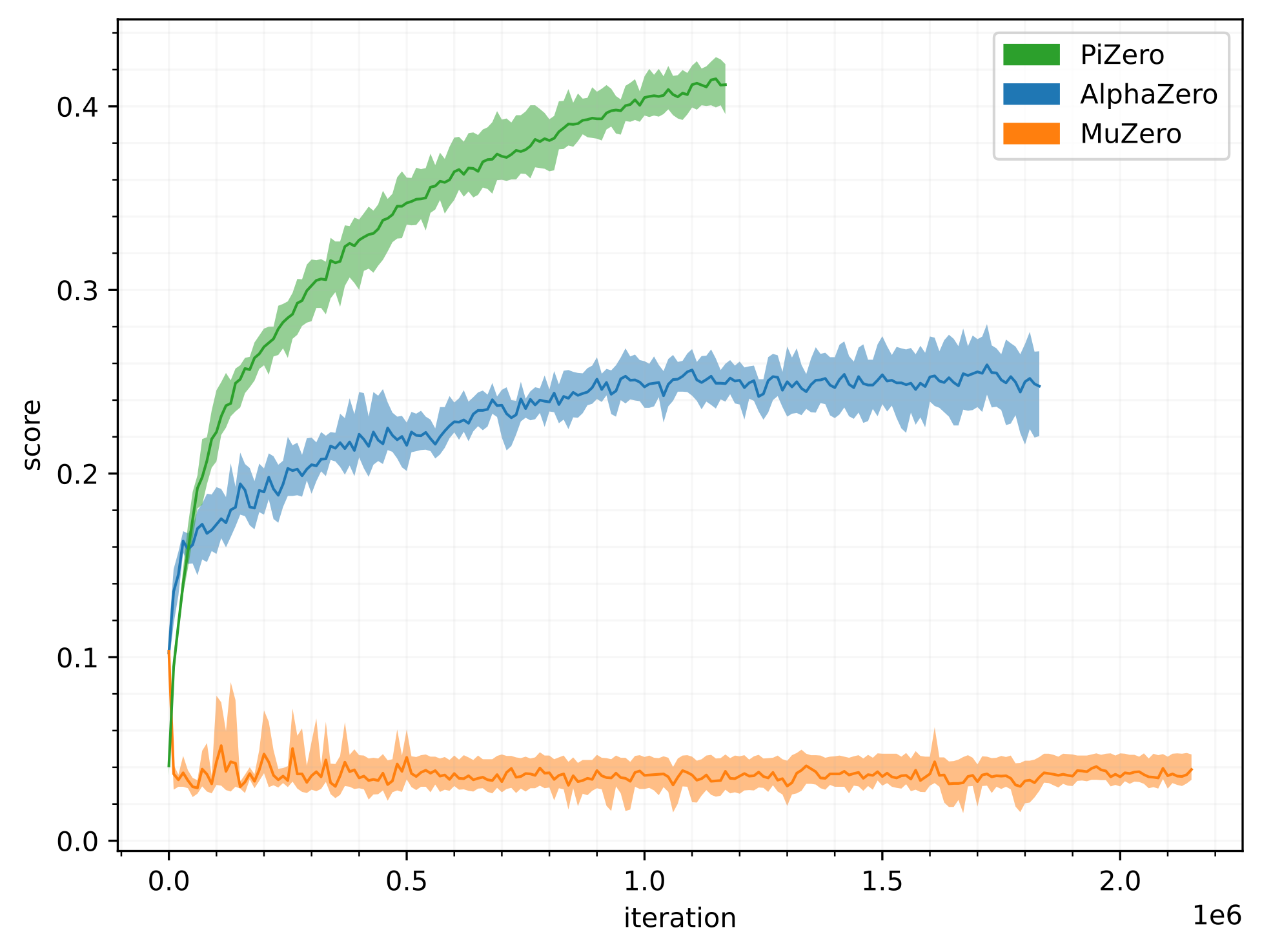}%
    \includegraphics[width=.333\linewidth]{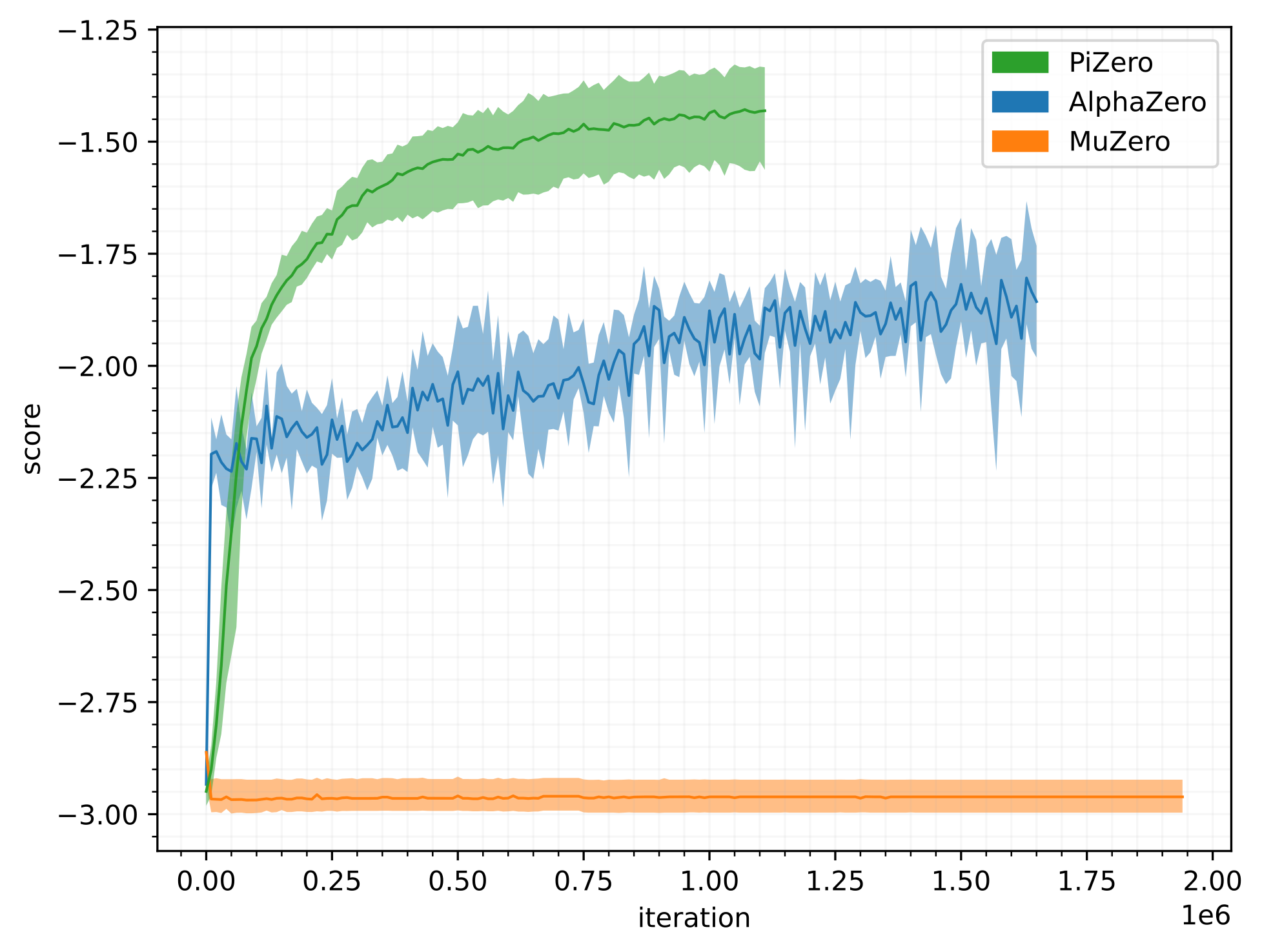}
    \includegraphics[width=.333\linewidth]{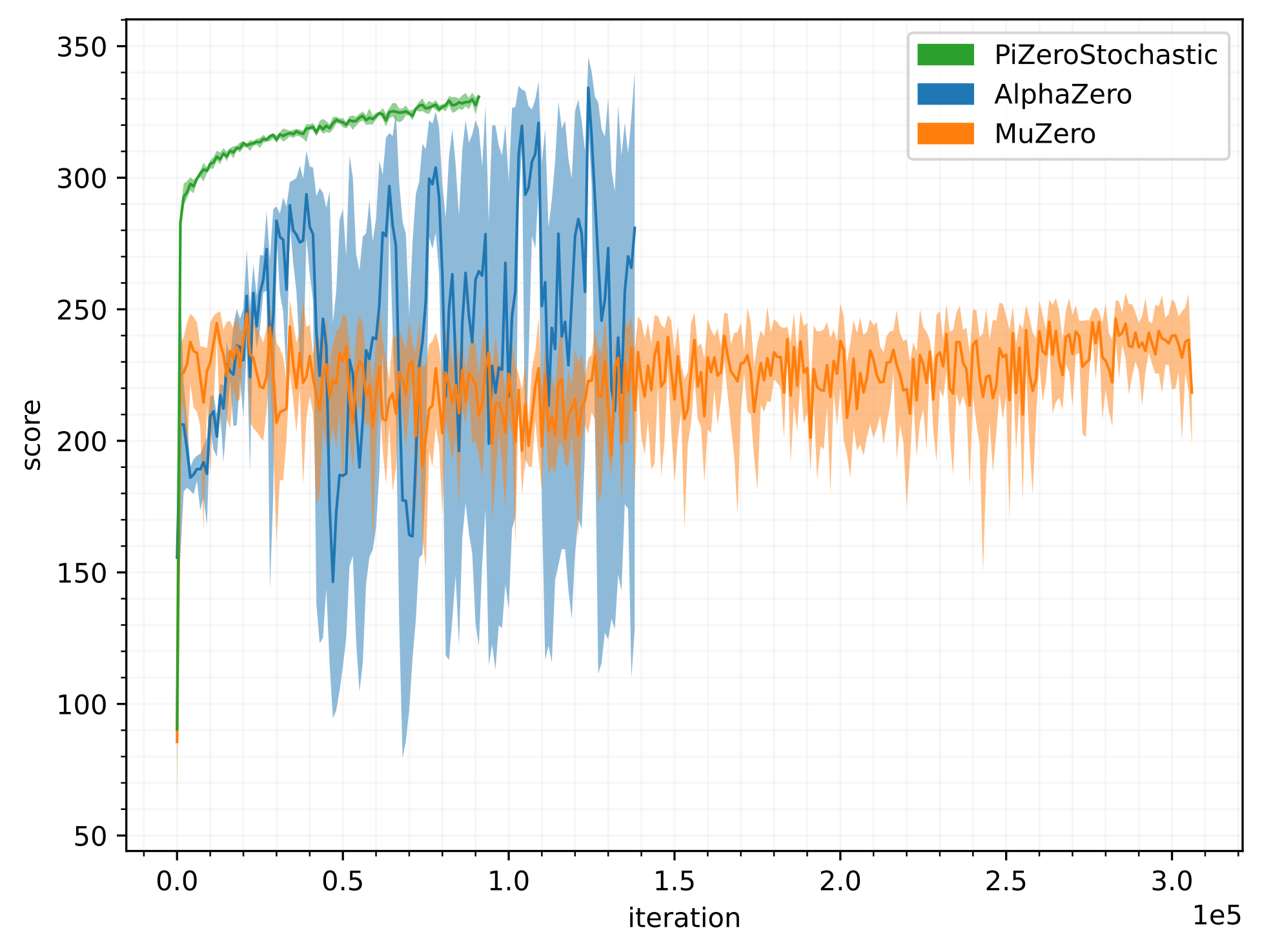}%
    \includegraphics[width=.333\linewidth]{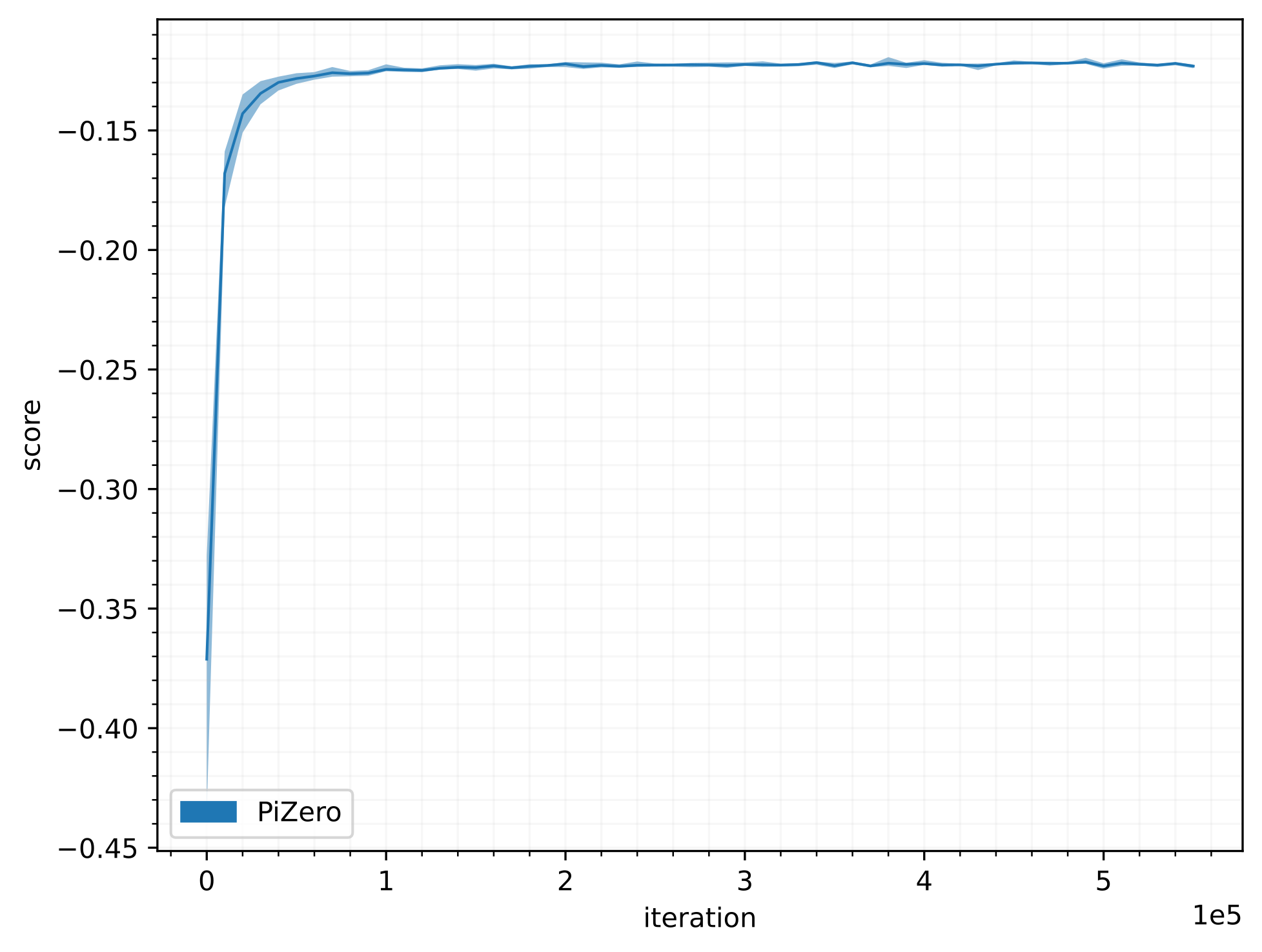}%
    \includegraphics[width=.333\linewidth]{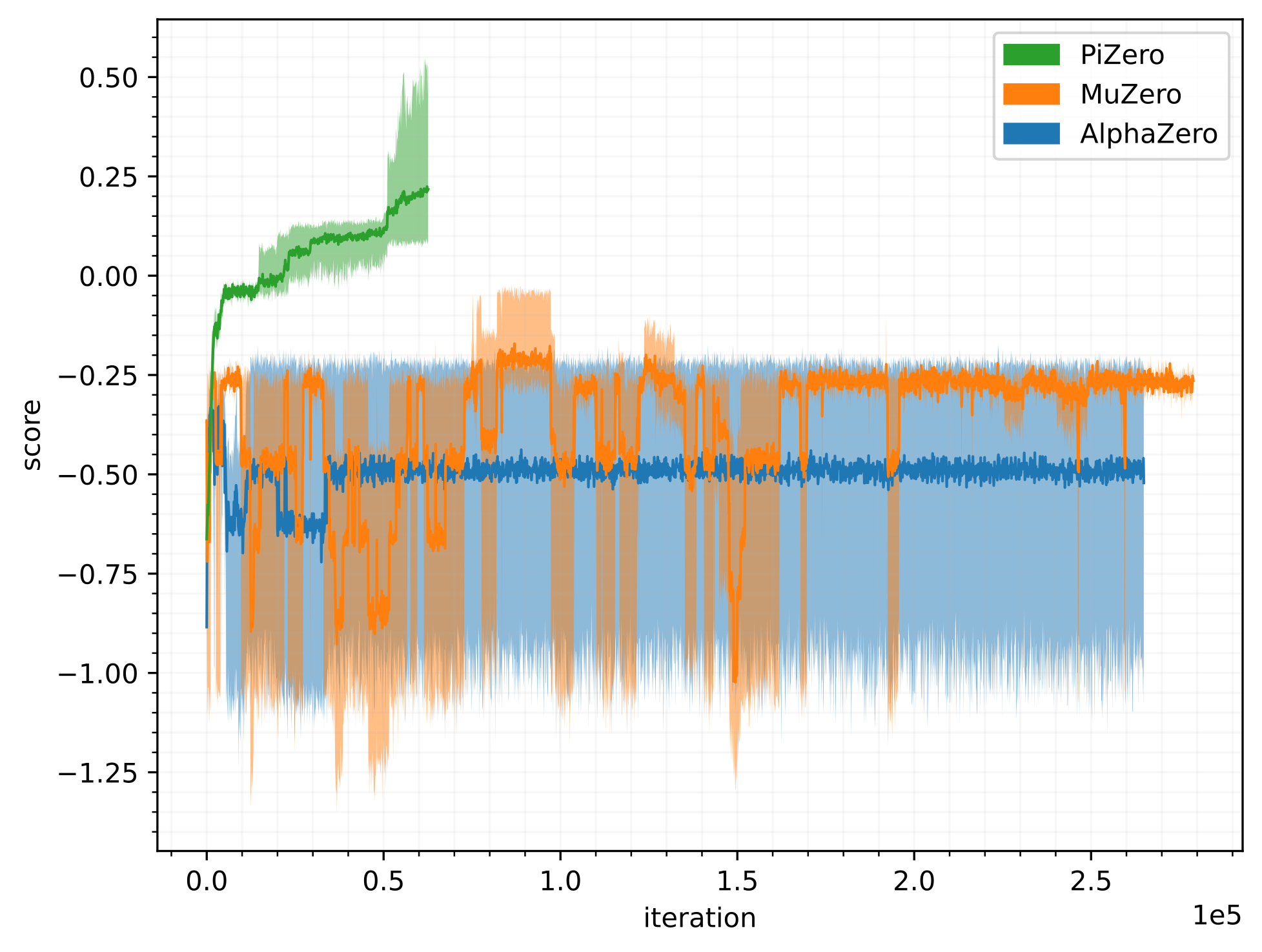}
    \caption{Results on TSP, Sokoban, Collect, 2048, FLP, and Pacman.}
    \label{fig:results}
\end{figure*}

An example state is shown in Figure \ref{fig:states}.
Results are shown in Figure~\ref{fig:results}.

\section{Conclusions and future research}
\label{sec:conclusion}

In this paper, we introduced a new method, called PiZero, that gives an agent the ability to plan in an abstract search space of its own creation that is completely decoupled from the real environment.
Unlike prior approaches, it allows the agent to perform high-level planning at arbitrary timescales and reason in terms of compound or temporally-extended actions.
In addition, the method is more general than comparable prior methods because it is able to handle settings with continuous action spaces and partial observability. 
In experiments on multiple domains, it outperformed comparable prior methods access to an environment simulator at execution time.

In the future, we would like to also apply the method to \emph{multiagent} settings where high-level planning is useful.
Such settings include \emph{real-time strategy (RTS)} games like MicroRTS~\citep{microrts}, ELF Mini-RTS~\citep{tian2017elf}, and Starcraft II~\citep{vinyals2017starcraft,samvelyan2019starcraft,vinyals2019grandmaster}.
Indeed, RTS games often contain various optimization problems as subtasks, like TSP (moving units around the map), FLP (deciding how to place units or buildings), and scheduling (management of resource production and consumption).

We would also like to investigate extending our method to \emph{partially-observable} multiagent settings, that is, imperfect-information games. One possible approach would be to replace MCTS in our method with an algorithm for solving imperfect-information games, such as \emph{counterfactual regret minimization (CFR)}~\citep{Zinkevich07:Regret}, its fastest modern variants \emph{discounted CFR}~\citep{Brown19:Solving} and \emph{predictive CFR+}~\citep{Farina21:Faster}, techniques that use deep learning for state generalization~\citep{Bowling15:Heads,pmlr-v97-brown19b,McAleer23:ESCHER}, or \emph{Fictitious Self-Play (FSP)}~\citep{Heinrich_2015,Heinrich16:Deep}. Further techniques for speeding up the planning inside imperfect-information games could potentially also be used, such as subgame solving~\citep{Brown17:Safe,Moravvcik17:DeepStacka,Brown18:Superhuman}, depth-limited subgame solving~\citep{Brown18:Depth,Brown19:Superhuman}, ReBeL~\citep{rebel}, and \emph{Student of Games (SoG)}~\citep{schmid2023student}. SoG uses an algorithm called \emph{growing-tree CFR (GT-CFR)}, which expands a tree asymmetrically toward the most relevant future states while iteratively refining values and policies.

There are also challenges that need to be addressed in order to scale this approach to more complex environments. 
The environment may have complex action and observation spaces---including high-dimensional arrays like images---that may require more sophisticated network architectures. These include convolutional networks~\citep{LeCun_1988, LeCun_1989}, pointer networks~\citep{Vinyals_2015}, transformers~\citep{Vaswani_2017}, and scatter connections that integrate spatial and non-spatial information~\citep{Vinyals19:Grandmaster}.

One may be able to get further speedups by using other ES algorithms instead~\citep{pmlr-v97-maheswaranathan19a,NEURIPS2020_a878dbeb} and by reducing variance of the smoothed gradient estimator using variance reduction techniques such as importance sampling (which samples from a different distribution than that being optimized) and control variates (which add random variates with zero mean, leaving the expectation unchanged).

Finally, some environments have very sparse rewards. This makes it difficult for the agent to learn since it receives little feedback from episode scores, especially in the initial stage of training when it acts randomly. In those cases, more sophisticated exploration techniques could be used, including approaches that seek novel observations, reward diverse behaviors, and create internal subgoals for the agent~\citep{houillon2013effect,aubret2019survey,ladosz2022exploration}.

\section{Acknowledgements}

This material is based on work supported by the Vannevar Bush Faculty Fellowship ONR N00014-23-1-2876, National Science Foundation grants IIS-1901403, CCF-1733556, RI-2312342, and RI-1901403, ARO award W911NF2210266, and NIH award A240108S001.



\begin{appendices}
\section{Additional related work}

Value Iteration Network (VIN)~\citep{vin} is a fully differentiable network with a planning module embedded within. It can learn to plan and predict outcomes that involve planning-based reasoning, such as policies for reinforcement learning. It uses a differentiable approximation of the value-iteration algorithm, which can be represented as a convolutional network, and is trained end-to-end using standard backpropagation.

Predictron~\citep{predictron} consists of a fully abstract model, represented by a Markov reward process, that can be rolled forward multiple ``imagined'' planning steps.
Each forward pass accumulates internal rewards and values over multiple planning depths.
The model is trained end-to-end so as to make these accumulated values accurately approximate the true value function.

Value Prediction Network (VPN)~\citep{vpn} integrates model-free and model-based RL methods into a single network. In contrast to previous model-based methods, it learns a dynamics model with abstract states that is trained to make action-conditional predictions of future returns rather than future observations.
VIN performs value iteration over the entire state space, which requires that 1) the state space is small and representable as a vector with each dimension corresponding to a separate state and 2) the states have a topology with local transition dynamics (such as a 2D grid). VPN does not have these limitations.
VPN is trained to make its predicted values, rewards, and discounts match up with those of the real environment~\citep[\S 3.3]{vpn}. In contrast, we do not use any kind of supervised training against the real environment's values, rewards, and discounts. We only seek to optimize the final episode score with respect to the policy's parameters.

Imagination-Augmented Agent (I2A)~\citep{racaniere2017imagination} augments a model-free agent with imagination by using environment models to simulate imagined trajectories, which are provided as additional context to a policy network. An environment model is any recurrent architecture which can be trained in an unsupervised fashion from agent trajectories: given a past state and current action, the environment model predicts the next state and observation. The imagined trajectory is initialized with the current observation and rolled out multiple time steps into the future by feeding simulated observations.

TreeQN~\citep{treeqn} is an end-to-end differentiable architecture that substitutes value function networks in discrete-action domains. 
Instead of directly estimating the state-action value from the current encoded state, as in Deep Q-Networks (DQN)~\citep{dqn}, it uses a learned dynamics model to perform planning up to some fixed-depth. The result is a recursive, tree-structured network between the encoded state and the predicted state-action values at the leafs. 
The authors also propose ATreeC, an actor-critic variant that augments TreeQN with a softmax layer to form a stochastic policy network.
Unlike our method, TreeQN/ATreeC performs limited-depth search rather than MCTS, which can grow the search tree asymmetrically and focus on more promising paths, thus scaling better.

MCTSnet~\citep{mcts_nets} incorporates simulation-based search inside a neural network, by expanding, evaluating and backing-up a vector embedding.
The parameters of the network are trained end-to-end using gradient-based optimisation.
When applied to small searches in the well-known planning problem Sokoban, it significantly outperformed MCTS baselines.

Aleph*~\citep{aleph_star} is a model-based reinforcement learning algorithm that combines A* search with a heuristic represented by a deep neural network.
The weights are learned through reinforcement learning: interacting with a simulated environment by performing actions and earning rewards.
State transitions are kept in a tree structure, and action
values are backpropagated along the tree to satisfy a time-difference equation.

\citet{guez2019investigation} propose that an entirely model-free approach, without special structure beyond standard neural network components such as convolutional networks and LSTMs, and without any strong inductive bias toward planning, can learn to exhibit many of the characteristics typically associated with a model-based planner.

Dreamer~\citep{Hafner2020Dream} is a reinforcement learning agent that solves long-horizon tasks from images purely by latent imagination. It efficiently learns behaviors by propagating analytic gradients of learned state values back through trajectories imagined in the compact state space of a learned world model.

Neural A*~\citep{neural_astar} is a data-driven search method for path planning problems.
It reformulates a canonical A* search algorithm to be differentiable and couples it with a convolutional encoder to form an end-to-end trainable neural network planner.
It solves a path planning problem by encoding a problem instance to a guidance map and then performing the differentiable A* search with the guidance map.
By learning to match the search results with ground-truth paths provided by experts, Neural A* can produce a path consistent with the ground truth accurately and efficiently.

\citet{ye2021mastering} proposed a sample efficient model-based visual RL algorithm built on MuZero, called EfficientZero.
\citet{schrittwieser2021online} introduce MuZero Unplugged, which combines MuZero with Reanalyze, an algorithm which uses model-based policy and value improvement operators to compute new improved training targets on existing data points, allowing efficient learning for data budgets.

Gumbel MuZero~\citep{gumbel_muzero} is a policy improvement algorithm based on sampling actions without replacement.
It replaces the more heuristic mechanisms by which AlphaZero selects and uses actions, both at root nodes and at non-root nodes.
It matches the state of the art on Go, chess, and Atari, and significantly improves prior performance when planning with few simulations.

\section{Experimental details}

The hyperparameters we used are shown in Table~\ref{tab:hyperparameters}.
We ran our experiments using the Slurm Workload Manager.
We used two nodes with 8 NVIDIA A100 SXM4 40 GB GPUs each.

\begin{table}
    \centering
    \begin{tabular}{|l|l|}
    \hline
    Hyperparameter & Value \\
    \hline
    Learning rate & \(10^{-3}\) \\
    Standard deviation & \(10^{-1}\) \\
    Batch size & \(10^3\) \\
    Trials & \(5\) \\
    Hidden layers & \(1\) \\
    Neurons per layer & \(64\) \\
    Abstract state dimension & \(64\) \\
    Abstract actions & \(4\) \\
    Abstract chance outcomes & \(3\) \\
    Planing loss unrolling steps & \(5\) \\
    Simulation budget & \(10\) \\
    \hline
    \end{tabular}
    \caption{Experimental hyperparameters.}
    \label{tab:hyperparameters}
\end{table}

\end{appendices}

\bibliography{dairefs,references}

\end{document}